\documentclass[journal]{IEEEtran}
\usepackage{hyperref} 
\usepackage{times}
\usepackage{epsfig}
\usepackage{graphicx}
\usepackage{amsmath}
\usepackage{amssymb}
\graphicspath{ {images/} }
\usepackage{colortbl}
\usepackage[table]{xcolor}
\usepackage{cite}
\usepackage{algorithm,algorithmic}
\usepackage{csquotes}
\usepackage{makecell}
\usepackage{multicol}
\usepackage{booktabs}
\usepackage{array,multirow}
\usepackage[caption=false]{subfig}

\begin{document}

\twocolumn
\normalsize

\title{Calorie Aware Automatic Meal Kit Generation from an Image}

\author{Ahmad~Babaeian~Jelodar,~and~Yu~Sun
\thanks{A. B. Jelodar, and Y. Sun are with the Department
of Computer Science and Engineering, University of South Florida,
FL, 33620 e-mail: ({ajelodar,yusun}@usf.edu).}
}


\maketitle


\begin{abstract}
Calorie and nutrition research has attained increased interest in recent years.
But, due to the complexity of the problem, literature in this area focuses on a limited subset of ingredients or dish types and simple convolutional neural networks or traditional machine learning.
Simultaneously, estimation of ingredient portions can help improve calorie estimation and meal re-production from a given image.
In this paper, given a single cooking image, a pipeline for calorie estimation and meal re-production for different servings of the meal is proposed.
The pipeline contains two stages.
In the first stage, a set of ingredients associated with the meal in the given image are predicted.
In the second stage, given image features and ingredients, portions of the ingredients and finally the total meal calorie are simultaneously estimated using a deep transformer based model.
Portion estimation introduced in the model helps improve the calorie estimation and is also beneficial for meal re-production in different serving sizes.
To demonstrate the benefits of the pipeline, the model can be used for meal kits generation.
To evaluate the pipeline, the large scale dataset Recipe1M is used.
Prior to experiments, the Recipe1M dataset is parsed and explicitly annotated with portions of ingredients.
Experiments show that using ingredients and their portions significantly improves calorie estimation.
Also, a visual interface is created in which a user can interact with the pipeline to reach accurate calorie estimations and generate a meal kit for cooking purposes.

\end{abstract}


%
\IEEEpeerreviewmaketitle

\section{Introduction}
\label{sec_intro}
Cooking related applications have become a popular research area in recent years spanning from tasks such as ingredient recognition \cite{inverse_cooking}, cooking motion recognition \cite{alibayev2021developing, alibayev2020estimating}, cooking activity understanding \cite{jelodar2018long}, dish classification \cite{food_recognition2, food_recognition3}, recipe generation from a single image \cite{recipe1m_1} to calorie estimation \cite{calorie1}, and recipe retrieval \cite{retrieval1}.
Food nutrition, and health are two
important aspects of our lives that require close monitoring and care and are strictly associated with cooking.
Specifically, the amount of calorie intake in a meal is an important matter of health.
Many research have addressed calorie estimations from a single image, but they only use simple small sized datasets with a few ingredients or dish types \cite{calorie1, calorie2, calorie3}.
They also lack simultaneous portion estimation of ingredients which can help improve calorie estimation, and reproduce the meal in different serving sizes.

\begin{figure} [!ht]
\centering
\includegraphics[width=8cm]{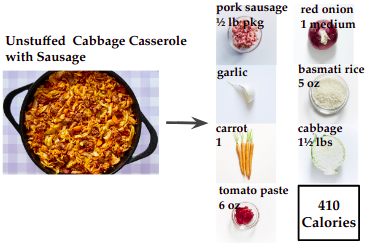}
\caption{An example of meal kits generation from a single image.}
\label{fig:introduction}
\end{figure}

Although ingredient recognition, and recipe generation from a single image have become growing areas of research in recent years, ingredient portion estimation is a neglected research area which includes very little literature that only focus on a small set of ingredients.
In this paper, we propose to predict ingredients, and estimate their portions while estimating the calorie of the given image.
For example, rather than only predicting the containing ingredients (e.g. carrot, cabbage, onion, etc) of a meal (e.g. cabbage casserole with sausage) or generating recipes, we focus on predicting ingredients and portions simultaneously (e.g. 1 carrot, 1.5 pounds cabbage, 1 onion) as shown in Figure \ref{fig:introduction}. 
One direct application of this research is the automatic generation of meal kits content.

Meal kit services, which offer and mail pre-portioned ingredients of specific meals have become very popular in recent years. These services also offer manually provided images and information about ingredients, and their portions.
Our proposed model can automatically extract knowledge about ingredients and portions from a given image making this work directly applicable to automatic meal kit generation.
To our knowledge there is no research that can automatically generate ingredients and portions at large scale and be used for applications such as automatic meal kit generation.
Visual information extraction from cooking images can be useful for many other applications such as calorie estimation, automatic recipe generation, and \textit{task graph generation} \cite{paulius2016functional, paulius2018functional} for automatic recipe preparation given a single image.

In this paper we propose a  two stage pipeline.
In the first stage, using a transformer based decoder \cite{transformer_2017}, the main and optional ingredients of a meal (illustrated in the given image) are generated sequentially.
In the second stage, all ingredients generated from the first stage are used for the task of total calorie estimation using a deep model with multiple encoder modules.
Three encoders are deployed in the model to model per ingredient calories, units, and portions and the total calorie intake of the image.
We finally introduce an application of this pipeline for meal kit generation.
This work has three main contributions:
\begin{itemize}
  \item Calorie estimation using ingredients and their portions  information from a single image.
  \item Portion estimation for the purpose of meal reproduction.
  \item Pipeline and interface for iterative meal kits generation.
\end{itemize}

\begin{figure*} [!ht]
\centering
\includegraphics[width=0.9\textwidth]{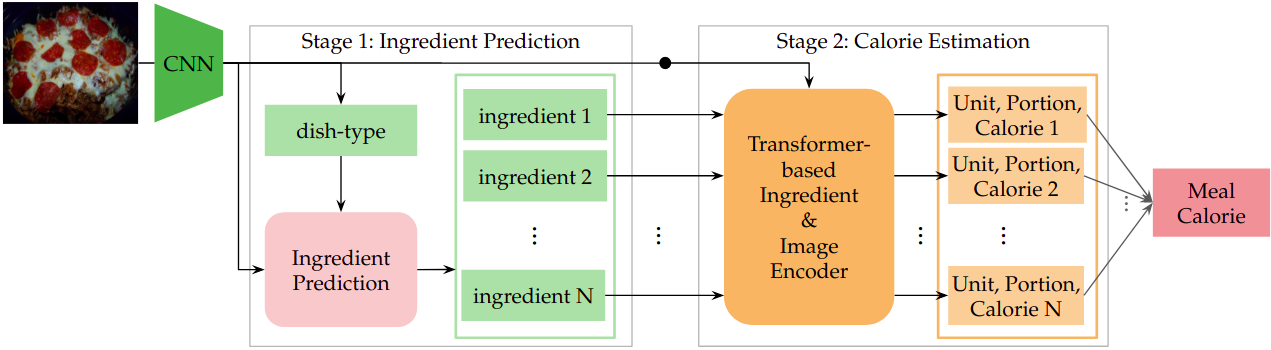}
\caption{The two stage pipeline for calorie estimation. Stage 1: Ingredient set generation. Stage 2: Estimation of meal calorie using intermediate estimates of ingredient portions, units, and calories.}
\label{fig:Pipeline}
\end{figure*}

The remainder of this paper is organized as follows:
In Section \ref{sec_related} we have an overview of the related work.
In Section \ref{sec_semiingredient} and \ref{sec_calorie} we discuss the first and second stages of the model respectively.
Finally, in Section \ref{sec_experiments} and \ref{sec_conclusion} we analyze the results and conclude the paper.

\section{Related Work} 
\label{sec_related}
\subsection{Dish Classification} \label{sub_rel_dish}

Dish (or food) classification can be considered as an application of image classification. 
Some work such as \cite{food_recognition2, food_recognition4} provide experimental studies on small scale datasets to recognize food (or dish) types from a single given image.
Many applications of dish classification incorporate non-visual context such as geo-location to increase dish classification accuracy \cite{food_recognition1, food_recognition5}.  
In \cite{food_recognition3}, to address the dynamic and changing nature of food and dish classification, Horiguchi et al. proposed a personalization based model for dish classification.
The commonality between the proposed work in this area is a small scale dataset and a deep learning model to address that.
We in this paper create our own set of dish types and use a state of the art deep learning network to perform dish classification.

\subsection{Ingredient Recognition} \label{sub_rel_ingredient}

Research in the area of ingredient recognition can be classified into two main categories: retrieval based, prediction.
In retrieval based applications, a list of ingredients or the whole recipe is retrieved based on creating an embedding and retrieving the appropriate image match from the dataset \cite{retrieval1, retrieval2, retrieval3, recipe1m_1}.
This body of work requires the predicted combination of ingredients to be a fixed set as seen in one of the datasets.
To handle this issue, ingredient prediction approaches inspired by multi-class modeling \cite{multi_label_classification1, multi_label_classification2, multi_label_classification3}, recurrent image captioning \cite{Review_ImageCaption1, Review_ImageCaption2, Review_ImageCaption3}, and auto-regressive list prediction methods emerged \cite{transformer_2017, inverse_cooking}.
Ingredient state recognition is also another field of study that has been under-studied.
Introducing new ingredients states datasets \cite{Ahmad_Paper, States_Transforms}, or addressing the states problem as image classification or multi-class labeling (i.e. ingredient-state tuple) problem \cite{Ahmad_Paper, icip_ahmad} are instances of research in this area.

One recent work that we use as our baseline for ingredient prediction is the inverse cooking research \cite{inverse_cooking}.
In this work, both ingredients and recipe are generated in auto-regressive manner using the transformer model \cite{inverse_cooking}.

\subsection{Portions Estimation} \label{sub_rel_portions}

Research on ingredient portions has mainly been conducted in the context of calorie estimation. 
In most of the work \cite{image_based1, image_based3, image_based4, image_based5} portions of ingredients (e.g. apple) are identified after image segmentation \cite{image_based1} and size computation \cite{image_based2} using various approaches (e.g. geometry, 3d modeling) for calorie estimation of very simple cooking images in small scale datasets \cite{image_based4}.
Also, approaching portions from a visual recognition and segmentation view may not be feasible in meals where the ingredients is not visually discerned (e.g. chicken in soup).
Therefore we approach the portions problem in a self-attention query based manner using a large scale dataset (i.e. Recipe1M).

\subsection{Calorie Estimation} \label{sub_rel_calorie}
Calorie estimation from image has gained attention in food and image processing research.
Some methods propose multi-stage pipelines to predict food categories/ingredients, identify portions/sizes and estimate the calorie intake of the food \cite{image_based1, image_based3,image_based4, image_based5}. 
Some of these methods take two input images to define depth and segmentation of food in the image \cite{image_based1, image_based4}.
These algorithms use model based or deep learning based methods for the recognition stage and standard nutritional fact tables
for calorie estimation \cite{image_based5}.
Some literature directly provide estimates of calorie from a food image \cite{direct_image1, direct_image2} by predicting the food category itself and directly mapping it to a calorie intake for that meal with \cite{direct_image2} or without a reference object.
The drawback to these methods is that they do not take into account the variety of ingredients different versions of a meal can have.
Some work propose a CNN-based direct image method that take into account multiple food in one image but still do not consider the containing ingredients of meals for calorie estimation \cite{direct_image3, calorie6, calorie7}.
Also, in \cite{calorie9} authors propose a food-estimation Bayesian framework for food-balance estimation which considers a limited number of food categories with a limited number of classes each with limited discrete values.

Most of the work done in the area of calorie estimation assumes that images are of food with clear segmented boundaries \cite{image_based1, image_based4, calorie3} and do not consider addressing more complex food such as mixed or cooked meals where the containing ingredients are not clear.
Another issue with these work is that the dataset used is very small and low diversity and the images are captured in a well-controlled setting.

On the other hand, there are a few literature that exploit ingredients for image-based calorie estimation \cite{calorie_ingredient1}.
In \cite{calorie_ingredient1} a deep learning based method is proposed for simultaneous learning of calories, categories, ingredients and cooking directions.
Datasets such as Japanese calorie-annotated food photo dataset and the American calorie-annotated food photo dataset \cite{calorie_ingredient1} are datasets with calorie annotations.

\section{Ingredient Generation} \label{sec_semiingredient} 

We  propose  a  two-stage  pipeline  for  calorie estimation of a given input image as shown in Figure \ref{fig:Pipeline}.
In the first stage (i.e. ingredient generation), given an image, the dish type, main ingredients and optional ingredients are predicted 
sequentially.
In the second stage, using the image and the generated ingredients, an estimate of portions and calories of each of the ingredients is provided.
Hierarchically the entire calorie intake of the given image is also estimated.
In this section, we will discuss the first stage and in the next section the second stage (i.e. portion and calorie estimation) is elaborated.

\begin{figure}[htp]
\centering
\subfloat[Predicting main ingredients iteratively (Main Ingredient Module).]{%
  \includegraphics[clip,width=0.8\columnwidth]{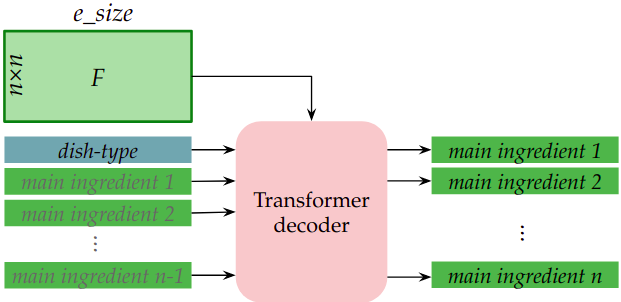}%
}

\subfloat[Predicting optional ingredients given main ingredients (optional Ingredient Module).]{%
  \includegraphics[clip,width=0.8\columnwidth]{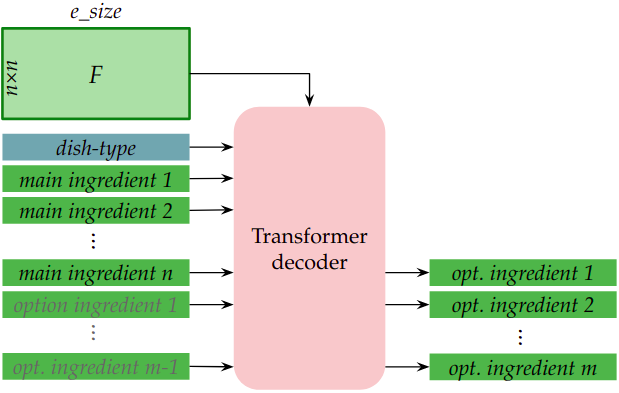}%
}
\caption{Prediction of ingredients given image features. A list of main ingredients are predicted in the first stage and afterwards a list of optional ingredients are predicted given the list of main ingredients. 
The list of main ingredients is predicted iteratively through iterations of correction.}
\label{fig:ingredient_prediction}
\end{figure}

\subsection{Dish Classification}
\label{sub_pipeline_dish}

The first stage of the pipeline starts with an estimation of the dish type \( d \), given the input image.
A convolutional neural network (with a Resnet base) is trained to classify the image into one of $N_d$ classes of dishes.
We name the predicted dish as $D$.
In the next steps, the predicted dish type is given along side the image as input to the ingredient generation model.
The dish vocabulary is also combined with the ingredient vocabulary to provide token embeddings for dish names alongside ingredient embeddings.

\subsection{Ingredient Generation}
\label{sub_details_ingredient}

The ingredient generation module utilizes a transformer decoder as in \cite{transformer_2017, inverse_cooking} and performs ingredient generation in two steps.
In the first step, the transformer generates a set of main ingredients, $I^{main}$, one step at a time as shown in the equation below and Figure \ref{fig:ingredient_prediction}.a.
The model provides confidences for each ingredient at each step that can be useful for correcting wrongful selections.

\begin{equation}
\label{equ_details_ingredient_main}
I^{main} = p(I^{main}_{j+1} / I^{main}_j, F, D)
\end{equation}

$I^{main}_j$ and $I^{main}_{j+1}$ are the j-th and (j+1)-th generated ingredients, $D$ is the predicted dish type from previous stage and $F$ is the image features extracted from the last convolutional layer of a pre-trained Resnet mapped to $(n\times n)\times e_{size}$ where $e$ is the embedding size.
All ingredient and dish tokens are projected to embeddings of size $e_{size}$.
The $p$ is a transformer decoder that takes as input previous ingredients and image features and generates ingredients one step at a time as shown in Figure \ref{fig:ingredient_prediction}.a.
In the second step, the transformer module generates a set of optional ingredients one step at a time given the main ingredients.

\begin{equation}
\label{equ_details_ingredient_non}
I^{opt} = p(I^{opt}_{k+1} / I^{opt}_k, I^{main}, F, D)
\end{equation}

All generated ingredients are merged and used in next stages of the pipeline $I = I^{opt}+I^{main}$.

\subsubsection{Formulation}
\label{sub_sub_details_ingredient}
The ingredient generation module takes image features and the predicted dish class, $D$, as input.
Image features are extracted from the final convolutional layer, $V \in \mathbb{R}^{M \times n \times n}$.
A $1 \times 1$ convolution layer and a reshape layer are applied to make the image features $F \in \mathbb{R}^{s_1 \times e_{size}}$.

\begin{equation}
\label{equ_image_features}
F = {reshape(conv_{1 \times 1}(V))}
\end{equation}

The module also takes as input a one-hot token matrix comprising of a dish token and ingredients.
Therefore, the vocabulary includes all ingredients in the dataset, $N_i$, combined with all dish vocabulary, $N_d$.
Therefore, the dish classes would be considered in the input vocabulary but not in the output vocabulary making the total vocabulary size $N=N_d+N_i$.
The input matrix of tokens is of size $I \in \mathbb{R}^{s_2 \times N}$ where $N$ is the size of vocabulary and $s_2$ is the number of maximum ingredients the model accepts as input.
The first token in $I$ is always the predicted dish class $D$ from the dish classification stage and the next tokens are generated ingredients from the previous step of the transformer.
The token matrix is projected to an embedding matrix, $E \in \mathbb{R}^{s_2 \times e_{size}}$, through an embedding layer.
The output of the transformer are generated ingredients at each step, 
$O_{ingr} \in \mathbb{R}^{s_2 \times N}$.

\section{Portion and Calorie Estimation} \label{sec_calorie} 

\begin{figure*} [!ht]
\centering
\includegraphics[width=0.8\textwidth]{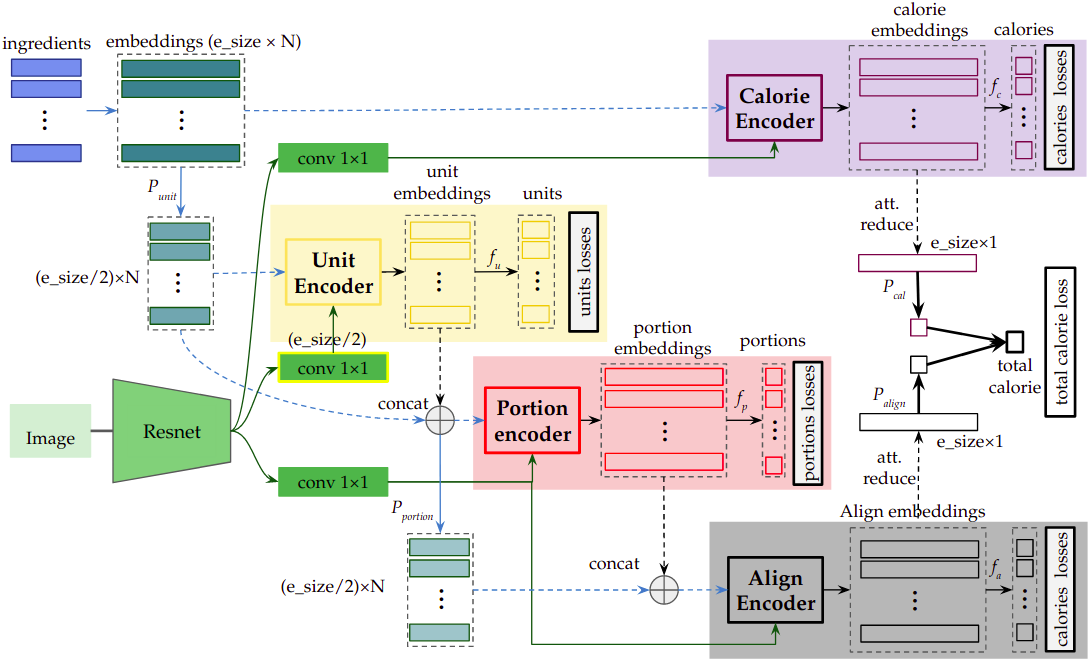}
\caption{The calorie estimation model with individual ingredient units, portions, and calories incorporation.}
\label{fig:calorie_estimation_module}
\end{figure*}

In the second stage, a two stream network is proposed to estimate the total calorie intake of a recipe (depicted in Figure \ref{fig:calorie_estimation_module}) given 
the generated ingredients and image features.
The first stream of the model is the calorie module that provides per ingredient calorie estimations.
The second stream includes a unit module, a portion module and an alignment module.
The unit module generates per ingredient unit predictions (e.g. teaspoon), the portion module uses the estimated units and the input ingredients to generate per ingredient portion estimations and the alignment module creates alignment between the generated units and portions using per ingredient calorie estimations.
The model is trained end-to-end.

\subsection{Inputs and Intermediate Outputs}
\label{sub_inputs_details_portions}

The proposed model takes as input all generated ingredients $I$ and image features $V$ and contains two streams with four modules and three intermediate outputs.
All encoders have two inputs; image features and ingredient embeddings.
Image features are extracted from the last convolutional layer, $V \in \mathbb{R}^{M \times n \times n}$,
and are projected and reshaped with a ($1 \times 1$) conv layer to $F_c \in \mathbb{R}^{s_1 \times e_{size}}$,
$F_u \in \mathbb{R}^{s_1 \times \frac{e_{size}}{2}}$,
$F_p \in \mathbb{R}^{s_1 \times e_{size}}$, and $F_a \in \mathbb{R}^{s_1 \times e_{size}}$ for the calorie encoder, unit encoder, portion encoder, and alignment encoder respectively.

All encoders create intermediate embeddings.
The calorie encoder, unit encoder, portion encoder, and alignment encoder create intermediate embeddings
$E_c \in \mathbb{R}^{s_2 \times e_{size}}$,
$E_u \in \mathbb{R}^{s_2 \times \frac{e_{size}}{2}}$,
$E_p \in \mathbb{R}^{s_2 \times e_{size}}$, and $E_a \in \mathbb{R}^{s_2 \times e_{size}}$
respectively.
The details of how these intermediate embeddings are generated are explained in Section \ref{sub_encoder_structure}.

Besides image features, each encoder has another set of inputs which is either ingredient embeddings or a combination of ingredient embeddings and intermediate embeddings from other encoders.
The generated ingredients from stage one are converted to ingredient embeddings through an embedding layer and are fed to the calorie encoder, $I \in \mathbb{R}^{s_2 \times e_{size}}$.
The original ingredient embeddings, $I$, are projected to smaller sized embeddings $I_u \in \mathbb{R}^{s_2 \times \frac{e_{size}}{2}}$ for the unit encoder.
The input for the portion encoder, $I_p \in \mathbb{R}^{s_2 \times e_{size}}$, and alignment encoder, $I_a \in \mathbb{R}^{s_2 \times {size}}$, are created by concatenating the smaller sized ingredient embeddings, $I_u$, and the intermediate unit embeddings, $E_u$.

\subsection{Encoder Structure}
\label{sub_encoder_structure}
Each of the four modules include an \textit{encoder} which follows the exact architecture of a transformer decoder \cite{transformer_2017}.
The transformer decoder has multiple identical transformer decoder layers and takes as input three sets of matrices; queries, keys and values.
Each of the transformer decoder layers in the encoder include two layers of multi-head attention layers with residual connections.
The general formulation of each of the multi-head attention layers is shown in Equation \ref{equ_unit_stream}.

\begin{equation}
\label{equ_unit_stream}
\begin{array}{l}
MultiHead(Q,K,V) = Concat(head_1, ..., head_h)W^o \\
where: head_i = Attention(QW_{i}^Q, KW_{i}^K, VW_{i}^V)
\end{array}
\end{equation}

where the projections are parameter matrices $W^{o} \in \mathbb{R}^{e_{size} \times s_2}$, $W_{i}^Q \in \mathbb{R}^{\frac{e_{size}}{h} \times s_2}$, $W_{i}^K \in \mathbb{R}^{\frac{e_{size}}{h} \times s_2}$,
$W_{i}^V \in \mathbb{R}^{\frac{e_{size}}{h} \times s_1} $ where $h$ is the number of heads, and $Attention$ is the scaled dot product attention mechanism from \cite{transformer_2017} and shown in Equation \ref{equ_sdpa}.

\begin{equation}
\label{equ_sdpa}
Attention(Q,K,V) = 
softmax(\frac{QK^T}{\sqrt{d_k}})V
\end{equation}

where $Q$ and $K \in \mathbb{R}^{\frac{e_{size}}{h} \times s_2}$, and $V \in \mathbb{R}^{\frac{e_{size}}{h} \times s_1}$ are projected matrices of queries, keys and values.
In our model both queries and keys are a set of ingredient embeddings and values are image features.
The last component of a transformer decoder layer is a position-wise feed forward network that is applied after the two multi-head attention layers \cite{transformer_2017}.
A stack of transformer decoder layers makes a transformer decoder.
The output of the transformer decoder is a matrix of embeddings $E \in \mathbb{R}^{s_2 \times e_{size}}$.
Each of the intermediate embeddings ($E_c, E_u, E_p, E_a$) discussed in \ref{sub_inputs_details_portions} are the output of a transformer decoder.
For more details on transformer decoders
readers are referred to \cite{transformer_2017}.

\subsection{The Final Network}
\label{sub_architecture}

The first stream of the network contains the calorie module and the second stream contains three modules (unit, portion and alignment encoders).
The modules contain encoders that create intermediate embeddings.
Calorie, portion, and alignment intermediate embeddings, $E_c$, $E_p$, $E_a$, $E_u$ are projected to intermediate outputs through (per ingredient) identical fully connected layers, $f_c$, $f_u$, $f_p$, $f_a$.

\begin{equation}
\label{equ_projected}
\begin{array}{l}
o_c = f_c(E_c) \\
O_u = f_u(E_u) \\
o_p = f_p(E_p) \\
o_a = f_a(E_a) 
\end{array}
\end{equation}

where $o_c \in \mathbb{R}^{s_2 \times 1}$, $O_u \in \mathbb{R}^{s_2 \times N_{units}}$, $o_p \in \mathbb{R}^{s_2 \times 1}$, and $o_a \in \mathbb{R}^{s_2 \times 1}$ are intermediate calorie, unit, portion, and alignment outputs for $s_2$ ingredient inputs.

The combination of intermediate unit and portion outputs provides a representation of the amount of each ingredient.
The purpose of adding an \textit{alignment module} with the ingredient calories loss incorporated is to align unit and portions with the calorie values. 

Attention reduction layers are applied to the calorie embeddings, $E_c$, from the first stream and alignment embeddings $E_a$ from the second stream to create a reduced calorie vector $r_c \in \mathbb{R}^{1 \times e_{size}}$ and a reduced alignment vector $r_a \in \mathbb{R}^{1 \times e_{size}}$ respectively. 
Projections $P_{cal}$ and $P_{align}$ are applied to $r_c$ and $r_a$ respectively and their outputs are concatenated to produce the output of the model (total calorie estimate). 
This two stream (four modules) model is named $T_{upc}$ henceforth.

\subsection{Losses}
\label{sub_losses}
Two types of loss are used in the entire model in five different locations.
One MSE loss is used for ingredient calorie estimation in the first stream ($L_1$).
The weighted cross-entropy loss is used for ingredient unit classification ($L_2$).
Units with less frequency in the training data are assigned a larger weight for loss computation.
Three MSE losses are used for portion estimation ($L_3$), calorie estimations in the alignment module ($L_4$) and total calorie estimation ($L_5$).
The final loss is computed as below with 
$\lambda_{1}$, $\lambda_{2}$, $\lambda_{3}$, $\lambda_{4}$, $\lambda_{5}$ being hyperparameters.

\begin{equation}
\label{equ_unit_stream}
\begin{array}{l}
L = \lambda_{1}L_1+\lambda_{2}L_2+\lambda_{3}L_3+\lambda_{4}L_4+\lambda_{5}L_5
\end{array}
\end{equation}

\section{Experiments and Results} \label{sec_experiments}

This section comprises of an overview of the dataset used in experiments, some of the implementation details, and all the experiments to evaluate the proposed model.

\subsection{Dataset} \label{sub_experiment_dataset}
We train and evaluate our models on the Recipe1M
dataset \cite{recipe1m_1}, composed of 1,029,720 recipes scraped from cooking websites. 
The dataset contains 720,639 training,
155,036 validation and 154,045 test recipes, with a title, a list of ingredients, a list of cooking instructions and/or an image. 
In our experiments, we use only
the recipes containing images, and remove recipes with less than 2 ingredients resulting in 252,547
training, 54,255 validation and 54,506 test samples \cite{inverse_cooking}.

Because the data was extracted by scraping cooking websites they are unstructured and include redundant ingredients.
We follow all the operations in \cite{inverse_cooking} (e.g. cluster 400 different cheese categories into one) and therefore reduce the number of ingredients from 16,823 to 1488.
Some of the ingredients in \cite{inverse_cooking} were clustered or split inaccurately. 
We performed a semi-automatic correction of some of the inaccurate clusters in \cite{inverse_cooking}.
For example we separated tomato from tomato sauce which were originally merged, or we merged sausages that were classified as separate classes with their brand names into one category.
We only maintain high frequency ingredients (top 95\%) which results in 202 ingredients.
After this stage we maintain 132,442 train and validation recipes and 23,602 test recipes.
We further automatically cluster recipes into 32 classes (i.e. dish names) using their ingredients list and recipe titles and remove outliers to keep 67,359 train and validation recipes and 11,743 test recipes.
We extract ingredients (e.g. tomato paste) and their portions (e.g. 1 spoon) from the provided text for each ingredient in the dataset \cite{recipe1m_1} and remove the recipes that contain extreme portion outliers or include ingredients with missing portions.
The final dataset contains 42,455 train and validation recipes and 7,575 test recipes, with a title, a list of (ingredients, portions, units) tuples, and images.

\subsection{Implementation Details} \label{sub_experiment_implementation}

Images were resized to 256 pixels in their shortest side and random crops of $224\times224$ were taken for training.
For evaluation the central $224\times224$ pixels were select.
For the transformer encoders, we use a transformer with 2 blocks and 8 multi-head attentions, each one with dimensionality 64.
For the last layer of the transformer, reduced embeddings of sizes 512, and 1024 were used.
To obtain image embeddings we use the last convolutional layer of ResNet-50 model which would be of size $2048\times7\times7$ (i.e. $M=2048, n=7$ in Section \ref{sec_semiingredient}).
All the word embeddings and all transformer decoder input vectors were set to 1024.
A maximum of 10 ingredients is used for each recipe. 
The models are trained with the Adam optimizer \cite{Adam} for 60 epochs.
Loss hyperparameters are all set to 1 with the exception of $\lambda_4$ being set to 0.1.
All parts of the model are implemented with PyTorch.
A GUI is implemented for user correction using Python and the Flask API.

\subsection{Results} \label{sub_results}

We perform ingredient generation in the first step of the pipeline.
State of the art ingredient generation models \cite{inverse_cooking} require much more improvements to be applicable to real world problems. 
Therefore, we proposed a multi-level model for ingredient generation.
We also include semi-automatic user correction at each level to provide accurate ingredients for the second stage.
For this, we created an interface using Flask and obtain user feedback on our results at each level of the ingredient generation stage \footnote{\href{http://www.rpal-eve.cee.usf.edu/}{http://www.rpal-eve.cee.usf.edu/}}.

\subsubsection{Ingredient Generation} \label{sub_experiment_ingredient}

In the first step of the experiments we implemented a Resnet based CNN for dish classification and achieved near to 50\% accuracy in classification of 32 dish types (e.g. omelette, cake, pizza, salad, etc).
Ingredient generation is performed as suggested in Section \ref{sec_semiingredient}.
The ingredient vocabulary set comprises of 202 ingredients (e.g. butter, chicken, strawberry, flour, etc).
The models for the ingredient generation include the dish types in their input vocabulary (i.e. 234 input tokens).
Table \ref{table_experiment_ingredient_1} shows intersection over union (IoU) results for three different experiments with or without dish types given as input.
Results for a) main ingredients generation: estimation of up to five main ingredients, b) optional ingredients generation: estimation of up to 10 optional ingredients given an image and main ingredients and c) all ingredients generation is shown in the table.
The results clearly show that when the model contains verbal knowledge about the dish type it estimates a more accurate ingredient list.

\begin{table} [!ht]
\centering
\caption{Intersection over union (IoU) for a) main ingredients, b) optional ingredients, and c) all ingredients generation.}

\label{table_experiment_ingredient_1}
 \begin{tabular}{|c |c |c |}
 \hline
\textbf{} & \textbf{Given dish} & \textbf{No dish given}\\ [0pt]
 \hline
 Main Ingredients  & 32.2\% & 27.3\% \\ [0pt]
 \hline
 Optional Ingredients  & 49.2\% & 47.7\% \\ [0pt]
  \hline
 All Ingredients  & 34.1\% & 31.5\% \\ [0pt]
 \hline
\end{tabular}
\end{table}

The ingredient estimates from the model for main ingredient generation is revised using user feedback.
Table \ref{table_main_ingrs} shows results for estimating one ingredient at a time given the previous ingredients are revised by the user.
As it can be observed in this table, the accuracy of generating main ingredients is much higher when revision happens.
To evaluate revision accuracy, the revised ingredient is fed back to the model (and not the actual ground-truth ingredient at that time step).
Therefore, the number of ingredients available at each time step in the test set is different for when the dish name is given in comparison to when it is not given as input.
Furthermore, having the dish name as input for the model improves performance.

\begin{table} [!ht]
\centering
\caption{Main Ingredient Prediction (Given N main ingredients)}
\label{table_main_ingrs}
 \begin{tabular}{| c | c | c |}
  \hline
\multirow{2}{*}{\textbf{Predicting}} & \multicolumn{2}{c|}{\textbf{Dish name}} \\ [0pt]
\cline{2-3}
& \textbf{Given} &  \textbf{Not given}\\ [0pt]
 \hline
 1st Ingredient  & 67.0\% & 56.8\% \\ [0pt]
 \hline
 2nd Ingredient  & 55.3\% & 49.1\% \\ [0pt]
 \hline
 3rd Ingredient  & 52.4\% & 44.4\% \\ [0pt]
 \hline
 4th Ingredient  & 45.3\% & 37.1\% \\ [0pt]
 \hline
 5th Ingredient  & 9.2\% & 21.9\% \\ [0pt]  
 \hline
\end{tabular}
\end{table}

\subsubsection{Portions and Units Estimation (meal kits)} \label{sub_experiment_portions}

\begin{figure*} [!ht]
\centering
\includegraphics[width=0.9\textwidth]{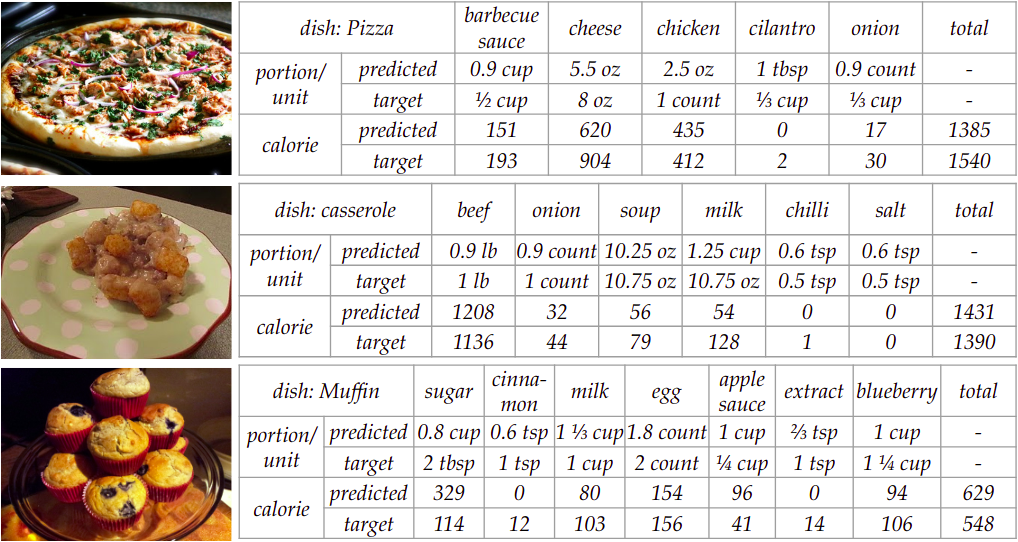}
\caption{Examples of results from different parts of the end-to-end pipeline which includes predicted dish name, generated ingredients, portions and calorie estimates.
The total predicted calorie intake and its ground-truth value is shown in the last column (gt: ground-truth).}
\label{fig:examples}
\end{figure*}

The portion encoder creates estimates of portions of the listed ingredients by providing a value and a unit.
The portions and units of an ingredient can be used to generate automatic meal kits procedures for a given image.
This stage of the model is evaluated based on the mean absolute error distance between the target and predicted portion and the accuracy of classification of units.
In our experiments six different units as shown in Table \ref{table_Units_Portions} is used.
In Table \ref{table_Units_Portions}, MAEs for predicted portions for each individual unit and the prior MAE for each unit is shown.
It can be observed that the calorie estimations are much better than the prior but the portions are not as good. 
The reasoning is that to have an accurate measurement of the amount of the ingredient an accurate combination of portion and unit is needed.
The unit estimation accuracy of the model is 72.3\%. 
In Figure \ref{fig:examples}, a few visual results of generated ingredients and their portions and calorie intakes for meal kits generation purposes is depicted.

\begin{table} [!ht]
\centering
\caption{MAE portion estimation of different units}
\label{table_Units_Portions}
 \begin{tabular}{| c | c | c |}
  \hline
\multirow{2}{*}{\textbf{Metric}} & \multicolumn{2}{c|}{\textbf{Calorie MAE}} \\ [0pt]
\cline{2-3}
& \textbf{Prior} &  \textbf{Estimated}\\ [0pt]
  \hline
 pound  & 323.7 & 162.8 \\ [0pt]
 \hline
 ounce  & 238.7 & 155.1 \\ [0pt]
 \hline
 cup  & 201.6 & 92.9 \\ [0pt]
 \hline
 count & 144.6 & 52.5 \\ [0pt]
 \hline
 tblsp & 99.4 & 68.9 \\ [0pt]  
 \hline
 tsp & 7.6 & 5.3 \\ [0pt]  
 \hline
 total & 144.2 & 78.5 \\ [0pt]  
 \hline
\end{tabular}
\end{table}

\subsubsection{Calorie Estimation} \label{sub_experiment_calorie}

For calorie estimation, we evaluated the proposed model using mean absolute error ($MAE$) and percentage mean absolute error ($MAE_\%$) on unseen test set.
We compare the final model ($T_{upc}$) with a few variations of transformer based models.
In one of the models we remove the ingredient based calorie training and only train the model for total recipe calories ($T_{calorie}$).
In another version, we maintain the same base but we only individually generate calorie values for each ingredient and a final calorie of the entire recipe ($T_{calories}$).

We also compare the final model with a simple neural network for generating a calorie for each ingredient individually and adding them up to generate the final calorie ($NN_{calories}$) and a single ingredient based neural network based on portions, units and calories of each ingredient and adding them up to generate the final calorie ($NN_{upc}$).

We compare the model with a trained CNN on recipe calories ($CNN$), recipe calories estimated using prior calorie of each ingredient ($P_{imean}$), and recipe calories estimated estimated using prior dish names ($P_{dish}$).

\begin{table} [!ht]
\centering
\caption{Calorie Estimation}
\label{table_experiment_calorie}
 \begin{tabular}{| c | c | c |}
  \hline
\multirow{2}{*}{\textbf{Model}} & \multicolumn{2}{c|}{\textbf{Total Calorie MAE}} \\ [0pt]
\cline{2-3}
& \textbf{$MAE$} &  \textbf{$MAE_\%$}\\ [0pt]
 \hline
 $T_{upc}$  & 279.4 & 37.5\% \\ [0pt]
 \hline
 $T_{calorie}$  & 394.5 & 49.9\% \\ [0pt]
 \hline
 $T_{calories}$  & 283.5 & 38.1\% \\ [0pt]
 \hline
  \hline
 $NN_{upc}$  & 306.7 & 39.7\% \\ [0pt]
 \hline
 $NN_{calories}$  & 310 & 40.9\% \\ [0pt]
 \hline
 $CNN$  & 380 & 49.8\% \\ [0pt]
 \hline
 \hline
 $P_{imean}$  & 323.3 & 44.7\% \\ [0pt]
 \hline
 $P_{dish}$  & 407 & 52.3\% \\ [0pt]
 \hline
\end{tabular}
\end{table}

We can observe from Table \ref{table_experiment_calorie} that the transformer model which uses intermediate ingredient portion and calorie estimates performs the best in both predicting ingredient calorie estimation and recipe calorie estimation.
Using ingredient based neural nets performs good but because the self attention between different ingredients is not modeled it performs worse than transformer based models.
The CNN based model reaches to 49.8\% $MAE_\%$ for recipe calorie estimation showing the need for ingredient incorporation in this application.
Just using the dish name ($P_{dish}$) and only knowing the ingredients ($P_{imean}$) perform relatively worse than the proposed model in estimating meal and ingredient calorie.

\subsection{Discussion} \label{sub_experiment_discussion}
Most state-of-the art models focus on ingredient retrieval given an image \cite{retrieval1, retrieval2}.
The models that generate ingredients (non-retrieval) from a given image have very low performance \cite{inverse_cooking}. We incorporate a state-of-the-art method as base with a slightly modified (i.e. corrected) dataset and add semi-automatic correction into the model to enhance the ingredient generation model for meal kits generation.

Our model also includes a stage (in the overall end-to-end pipeline) where portions and units (e.g. 1 spoon of oil) are generated for each ingredient using both attention between ingredients themselves and their underlying image features using transformer properties.
An element in the portions generation stage that is unique to our model and beneficial is the containment of six unit types in the output which can automatically be mapped to meal kit content generation.

Another potential property of the model is the use of per ingredient unit and portion estimation to backtrack and provide re-computed ingredient amounts given a new serving amount of the meal.

To our knowledge, this work is the first work on large scale calorie estimation on images with intermediate ingredient and portions estimation where the ingredients may or may not be (e.g. salt) visually seen in the image.
Also, to prepare data for portions and calorie estimation we removed instances that lacked enough data from the Recipe1M dataset and therefore the dataset used in the experiments is uniquely tailored for this application making it impossible to compare the results with any baseline methods.  

\section{Conclusion and Future Work} \label{sec_conclusion} 

The main objective of this paper was calorie estimation and identifying the underlying ingredients and their portions for meal kits generation.
We proposed a pipeline using deep encoders to extract image features and generate ingredients (and their portions) from a given image.
The ingredients are extracted using an auto-regressive encoder which captures both relative association between ingredients through self attention and ingredients and the image features through the transformer and Resnet features.
The portions and furthermore calories of ingredients are also extracted with the assumption that knowing all ingredients contained in a recipe image can contribute to the portion knowledge base of a recipe (i.e. self-attention between portions of ingredients).
The total calorie is estimated using all generated mid-level knowledge in an end-to-end manner.
The current pipeline can semi-automatically generate ingredients with portions, and calories and a final calorie estimate of the entire recipe.
In future work, we plan to extend experiments in a wider range of ingredients and recipes where more accurate portion estimates is available. 
We also plan to integrate this pipeline with robotic manipulation and use the predicted ingredients and their portions, state changes to infer manipulation tools based on the relationships between ingredients and tools \cite{sun2014object,ren2013human}. Eventually the complete system will be able to infer a robot manipulation task graph for cooking the meal in an image from its meal kit \cite{sakib2021functional, sakib2021evaluating}.

\ifCLASSOPTIONcaptionsoff
  \newpage
\fi

{\small
\bibliographystyle{unsrt}
\bibliography{main}

\begin{thebibliography}{10}

\bibitem{inverse_cooking}
Amaia Salvador, Michal Drozdzal, Xavier Giro-i Nieto, and Adriana Romero.
\newblock Inverse cooking: Recipe generation from food images.
\newblock In {\em CVPR}, June 2019.

\bibitem{alibayev2021developing}
Maxat Alibayev, David Paulius, and Yu~Sun.
\newblock Developing motion code embedding for action recognition in videos.
\newblock In {\em 2020 25th International Conference on Pattern Recognition
  (ICPR)}, pages 7529--7536. IEEE, 2021.

\bibitem{alibayev2020estimating}
Maxat Alibayev, David Paulius, and Yu~Sun.
\newblock Estimating motion codes from demonstration videos.
\newblock In {\em 2020 IEEE/RSJ International Conference on Intelligent Robots
  and Systems (IROS)}, pages 4257--4262. IEEE, 2020.

\bibitem{jelodar2018long}
Ahmad~Babaeian Jelodar, David Paulius, and Yu~Sun.
\newblock Long activity video understanding using functional object-oriented
  network.
\newblock {\em IEEE Transactions on Multimedia}, 21(7):1813--1824, 2018.

\bibitem{food_recognition2}
Yi~Sen Ng, Wanqi Xue, Wei Wang, and Panpan Qi.
\newblock Convolutional neural networks for food image recognition: An
  experimental study.
\newblock In {\em International Workshop on Multimedia Assisted Dietary
  Management}, page 33–41, 2019.

\bibitem{food_recognition3}
Shota Horiguchi, Sosuke Amano, Makoto Ogawa, and Kiyoharu Aizawa.
\newblock Personalized classifier for food image recognition.
\newblock {\em IEEE Transactions on Multimedia}, 20:2836--2848, 2018.

\bibitem{recipe1m_1}
Javier Marin, Aritro Biswas, Ferda Ofli, Nicholas Hynes, Amaia Salvador, Yusuf
  Aytar, Ingmar Weber, and Antonio Torralba.
\newblock Recipe1m+: A dataset for learning cross-modal embeddings for cooking
  recipes and food images.
\newblock {\em PAMI}, 2019.

\bibitem{calorie1}
Pei-Yu Chi, Jen hao Chen, Hao-Hua Chu, and Jin-Ling Lo.
\newblock Enabling calorie-aware cooking in a smart kitchen.
\newblock In {\em PERSUASIVE}, volume 5033 of {\em Lecture Notes in Computer
  Science}, pages 116--127. Springer, 2008.

\bibitem{retrieval1}
Jing-jing Chen, Chong-Wah Ngo, and Tat-Seng Chua.
\newblock Cross-modal recipe retrieval with rich food attributes.
\newblock In {\em ACM International Conference on Multimedia}, MM ’17, page
  1771–1779, 2017.

\bibitem{calorie2}
{Wen Wu} and {Jie Yang}.
\newblock Fast food recognition from videos of eating for calorie estimation.
\newblock In {\em 2009 IEEE International Conference on Multimedia and Expo},
  pages 1210--1213, 2009.

\bibitem{calorie3}
P.~{Pouladzadeh}, G.~{Villalobos}, R.~{Almaghrabi}, and S.~{Shirmohammadi}.
\newblock A novel svm based food recognition method for calorie measurement
  applications.
\newblock In {\em 2012 IEEE International Conference on Multimedia and Expo
  Workshops}, pages 495--498, 2012.

\bibitem{paulius2016functional}
David Paulius, Yongqiang Huang, Roger Milton, William~D Buchanan, Jeanine Sam,
  and Yu~Sun.
\newblock Functional object-oriented network for manipulation learning.
\newblock In {\em 2016 IEEE/RSJ International Conference on Intelligent Robots
  and Systems (IROS)}, pages 2655--2662. IEEE, 2016.

\bibitem{paulius2018functional}
David Paulius, Ahmad~B Jelodar, and Yu~Sun.
\newblock Functional object-oriented network: Construction \& expansion.
\newblock In {\em 2018 IEEE International Conference on Robotics and Automation
  (ICRA)}, pages 5935--5941. IEEE, 2018.

\bibitem{transformer_2017}
Ashish Vaswani, Noam Shazeer, Niki Parmar, Jakob Uszkoreit, Llion Jones,
  Aidan~N Gomez, \L~ukasz Kaiser, and Illia Polosukhin.
\newblock Attention is all you need.
\newblock In {\em Advances in Neural Information Processing Systems 30}, pages
  5998--6008. 2017.

\bibitem{food_recognition4}
N.~{Martinel}, G.~L. {Foresti}, and C.~{Micheloni}.
\newblock Wide-slice residual networks for food recognition.
\newblock In {\em WACV}, pages 567--576, March 2018.

\bibitem{food_recognition1}
Luis Herranz, Shuqiang Jiang, and Ruihan Xu.
\newblock Modeling restaurant context for food recognition.
\newblock {\em IEEE Transactions on Multimedia}, 19(2):430–440, February
  2017.

\bibitem{food_recognition5}
R.~{Xu}, L.~{Herranz}, S.~{Jiang}, S.~{Wang}, X.~{Song}, and R.~{Jain}.
\newblock Geolocalized modeling for dish recognition.
\newblock {\em IEEE Transactions on Multimedia}, 17(8):1187--1199, Aug 2015.

\bibitem{retrieval2}
Micael Carvalho, R\'{e}mi Cad\`{e}ne, David Picard, Laure Soulier, Nicolas
  Thome, and Matthieu Cord.
\newblock Cross-modal retrieval in the cooking context: Learning semantic
  text-image embeddings.
\newblock In {\em The 41st International ACM SIGIR Conference on Research and
  Development in Information Retrieval}, SIGIR ’18, page 35–44, 2018.

\bibitem{retrieval3}
Xin Wang, Devinder Kumar, Nicolas Thome, Matthieu Cord, and Frederic Precioso.
\newblock Recipe recognition with large multimodal food dataset.
\newblock In {\em ICMEW}, 2015.

\bibitem{multi_label_classification1}
J.~Wang, Y.~Yang, J.~Mao, Z.~Huang, C.~Huang, and W.~Xu.
\newblock Cnn-rnn: A unified framework for multi-label image classification.
\newblock {\em CVPR}, 2016.

\bibitem{multi_label_classification2}
Jinseok Nam, Eneldo Loza~Menc\'{\i}a, Hyunwoo~J Kim, and Johannes
  F\"{u}rnkranz.
\newblock Maximizing subset accuracy with recurrent neural networks in
  multi-label classification.
\newblock In {\em NeurIPS}, pages 5413--5423. 2017.

\bibitem{multi_label_classification3}
Chih-Kuan Yeh, Wei-Chieh Wu, Wei-Jen Ko, and Yu-Chiang~Frank Wang.
\newblock Learning deep latent spaces for multi-label classification.
\newblock In {\em AAAI}, AAAI’17, page 2838–2844, 2017.

\bibitem{Review_ImageCaption1}
T.~Yao, Y.~Pan, Y.~Li, Z.~Qiu, and T.~Mei.
\newblock Boosting image captioning with attributes.
\newblock {\em ICCV}, 00:4904--4912, Oct. 2018.

\bibitem{Review_ImageCaption2}
Jeff Donahue, Lisa~Anne Hendricks, Sergio Guadarrama, Marcus Rohrbach,
  Subhashini Venugopalan, Kate Saenko, and Trevor Darrell.
\newblock Long-term recurrent convolutional networks for visual recognition and
  description.
\newblock {\em CoRR}, abs/1411.4389, 2014.

\bibitem{Review_ImageCaption3}
S.~Venugopalan, L.~A. Hendricks, M.~Rohrbach, R.~Mooney, T.~Darrell, and
  K.~Saenko.
\newblock Captioning images with diverse objects.
\newblock In {\em CVPR}, pages 1170--1178, July 2017.

\bibitem{Ahmad_Paper}
A.~B. {Jelodar}, M.~S. {Salekin}, and Y.~{Sun}.
\newblock Identifying object states in cooking-related images.
\newblock {\em arXiv preprint arXiv:1805.06956}, May 2018.

\bibitem{States_Transforms}
P.~Isola, J.~J. Lim, and E.~H. Adelson.
\newblock Discovering states and transformations in image collections.
\newblock {\em CVPR}, 2015.

\bibitem{icip_ahmad}
A.~B. {Jelodar} and Y.~{Sun}.
\newblock Joint object and state recognition using language knowledge.
\newblock In {\em 2019 IEEE International Conference on Image Processing
  (ICIP)}, pages 3352--3356, Sep. 2019.

\bibitem{image_based1}
P.~{Pouladzadeh}, S.~{Shirmohammadi}, and R.~{Al-Maghrabi}.
\newblock Measuring calorie and nutrition from food image.
\newblock {\em IEEE Transactions on Instrumentation and Measurement},
  63(8):1947--1956, 2014.

\bibitem{image_based3}
Austin Myers, Nick Johnston, Vivek Rathod, Anoop Korattikara, Alex Gorban,
  Nathan Silberman, Sergio Guadarrama, George Papandreou, Jonathan Huang, and
  Kevin Murphy.
\newblock Im2calories: towards an automated mobile vision food diary.
\newblock In {\em ICCV}, 2015.

\bibitem{image_based4}
P.~{Pouladzadeh}, S.~{Shirmohammadi}, and A.~{Yassine}.
\newblock Using graph cut segmentation for food calorie measurement.
\newblock In {\em 2014 IEEE International Symposium on Medical Measurements and
  Applications (MeMeA)}, pages 1--6, 2014.

\bibitem{image_based5}
Shaobo Fang, Fengqing Zhu, Chufan Jiang, Song Zhang, Carol~J. Boushey, and
  Edward~J. Delp.
\newblock A comparison of food portion size estimation using geometric models
  and depth images.
\newblock {\em 2016 IEEE International Conference on Image Processing (ICIP)},
  pages 26--30, 2016.

\bibitem{image_based2}
Hsin-Chen Chen, Wenyan Jia, Yaofeng Yue, Zhaoxin Li, Yung-Nien Sun, John~D.
  Fernstrom, and Mingui Sun.
\newblock Model-based measurement of food portion size for image-based dietary
  assessment using 3d/2d registration.
\newblock {\em Measurement science and technology}, 24 10, 2013.

\bibitem{direct_image1}
T.~{Miyazaki}, G.~C. {de Silva}, and K.~{Aizawa}.
\newblock Image-based calorie content estimation for dietary assessment.
\newblock In {\em 2011 IEEE International Symposium on Multimedia}, pages
  363--368, 2011.

\bibitem{direct_image2}
Koichi Okamoto and Keiji Yanai.
\newblock An automatic calorie estimation system of food images on a
  smartphone.
\newblock In {\em Proceedings of the 2nd International Workshop on Multimedia
  Assisted Dietary Management}, MADiMa ’16, page 63–70. Association for
  Computing Machinery, 2016.

\bibitem{direct_image3}
Wataru Shimoda and Keiji Yanai.
\newblock Cnn-based food image segmentation without pixel-wise annotation.
\newblock In Vittorio Murino, Enrico Puppo, Diego Sona, Marco Cristani, and
  Carlo Sansone, editors, {\em New Trends in Image Analysis and Processing --
  ICIAP 2015 Workshops}, pages 449--457. Springer International Publishing,
  2015.

\bibitem{calorie6}
T.~{Ege}, Y.~{Ando}, R.~{Tanno}, W.~{Shimoda}, and K.~{Yanai}.
\newblock Image-based estimation of real food size for accurate food calorie
  estimation.
\newblock In {\em 2019 IEEE Conference on Multimedia Information Processing and
  Retrieval (MIPR)}, pages 274--279, 2019.

\bibitem{calorie7}
Parisa Pouladzadeh and Shervin Shirmohammadi.
\newblock Mobile multi-food recognition using deep learning.
\newblock {\em ACM Trans. Multimedia Comput. Commun. Appl.}, 13(3s), August
  2017.

\bibitem{calorie9}
K.~{Aizawa}, Y.~{Maruyama}, H.~{Li}, and C.~{Morikawa}.
\newblock Food balance estimation by using personal dietary tendencies in a
  multimedia food log.
\newblock {\em IEEE Transactions on Multimedia}, 15(8):2176--2185, 2013.

\bibitem{calorie_ingredient1}
Takumi Ege and Keiji Yanai.
\newblock Image-based food calorie estimation using knowledge on food
  categories, ingredients and cooking directions.
\newblock In {\em Proceedings of the on Thematic Workshops of ACM Multimedia
  2017}, Thematic Workshops ’17, page 367–375, 2017.

\bibitem{Adam}
Diederik~P. Kingma and Jimmy Ba.
\newblock Adam: {A} method for stochastic optimization.
\newblock In Yoshua Bengio and Yann LeCun, editors, {\em 3rd International
  Conference on Learning Representations, {ICLR} 2015, San Diego, CA, USA, May
  7-9, 2015, Conference Track Proceedings}, 2015.

\bibitem{sun2014object}
Yu~Sun, Shaogang Ren, and Yun Lin.
\newblock Object--object interaction affordance learning.
\newblock {\em Robotics and Autonomous Systems}, 62(4):487--496, 2014.

\bibitem{ren2013human}
Shaogang Ren and Yu~Sun.
\newblock Human-object-object-interaction affordance.
\newblock In {\em 2013 IEEE Workshop on Robot Vision (WORV)}, pages 1--6. IEEE,
  2013.

\bibitem{sakib2021functional}
Md~Sakib, David Paulius, and Yu~Sun.
\newblock Functional task tree generation from a knowledge graph to solve
  unseen problems.
\newblock {\em arXiv preprint arXiv:2112.02433}, 2021.

\bibitem{sakib2021evaluating}
Md~Sadman Sakib, Hailey Baez, David Paulius, and Yu~Sun.
\newblock Evaluating recipes generated from functional object-oriented network.
\newblock {\em 19th International Conference on Ubiquitous Robots}, pages 1--4,
  2021.

\end{thebibliography}
}

\end{document}